# A Freely Available Wide Coverage Morphological Analyzer for English*

*Daniel Karp*[†], *Yves Schabes*, *Martin Zaidel*, and *Dania Egedi*
Department of Computer and Information Science
University of Pennsylvania
Philadelphia PA 19104-6389 USA
dkarp/schabes/zaidel/egedi@cis.upenn.edu

## Abstract

This paper presents a morphological lexicon for English that handle more than 317000 inflected forms derived from over 90000 stems. The lexicon is available in two formats. The first can be used by an implementation of a two-level processor for morphological analysis (Karttunen and Wittenburg, 1983; Antworth, 1990). The second, derived from the first one for efficiency reasons, consists of a disk-based database using a UNIX hash table facility (Seltzer and Yigit, 1991). We also built an X Window tool to facilitate the maintenance and browsing of the lexicon. The package is ready to be integrated into an natural language application such as a parser through hooks written in Lisp and C.

To our knowledge, this package is the only available free English morphological analyzer with very wide coverage.

## 1 Introduction

Morphological analysis has experienced great success since the introduction of two-level morphology (Koskenniemi, 1983; Karttunen, 1983). Two-level morphology and its implementation are now well understood both linguistically and computationally (Karttunen, 1983; Karttunen and Wittenburg, 1983; Koskenniemi, 1985; Barton et al., 1987; Koskenniemi and Church, 1988). This computational model has proved to be well suited for many languages. Although there are some proprietary wide coverage morphological analyzers for English, to our knowledge those that are freely available provide only very small coverage.

Working from the 1979 edition of the Collins Dictionary of the English Language available through ACL-DCI (Liberman, 1989), we constructed lexicons for PC-KIMMO (Antworth, 1990), a public domain implementation of a two-level processor. Using the morphological rules for English inflections provided by Karttunen and Wittenburg (1983) and our lexicons, PC-KIMMO outputs all possible analyses of each input word, giving its root form and its inflectional attributes. To improve performance, we used PC-KIMMO as a generator on our lexicons to build a disk-based hashed database with a UNIX database facility (Seltzer and Yigit, 1991). Both formats, PC-KIMMO and database, are now available for distribution. We also provide an X Window tool for the database to facilitate maintenance and access. Each format contains the morphological information for over 317000 English words. The morphological database for English runs under UNIX; PC-KIMMO runs under UNIX and on a PC.

This package can be easily embedded into a natural language parser; hooks for accessing the morphological database from a parser are provided for both Lucid Common Lisp and C. This morphological database is currently being used in a graphical workbench (XTAG) for the development of tree-adjoining grammars and their parsers (Paroubek et al., 1992).

## 2 Lexicons for PC-KIMMO

We used the set of morphological rules for English described by Karttunen and Wittenburg (1983). The rules handle the following phenomena (among others[1]): epenthesis, $y$ to $i$ correspondences, s-deletion, elision, $i$ to $y$ correspondences, gemination, and hyphenation. In addition to the set of rules, PC-KIMMO requires lexicons. We derived PC-KIMMO-style lexicons from the 1979 edition of the Collins Dictionary of the English Language. The 90000-odd roots[2] in the lexicon yield over 317000 inflected forms.

The lexicons use the following parts of speech: verbs (V), pronoun (Pron), preposition (Prep), noun (N), determiner (D), conjunction (Conj), adverb (Adv), and adjective (A). Figure 1 shows the distribution of these parts of speech in the two formats: The first column is the distribution of the root forms in the PC-KIMMO lexicon files, and the second column is the distribution for the inflected forms derived from the lexicons and stored in the database. For each word, the lexicon lists its lexical form, a continuation class, and a parse. The continuation class specifies which inflections the lexical form can undergo. At most, a noun root engenders four inflections (singular, plural, singular genitive, plural genitive); an adjective root, three (base, com-parative, superlative); and a verb root, five (infinitive, third-person singular present, simple past, past participle, progressive). The exact number generated by any given root depends on its continuation class.

---

*This work was partially supported by DARPA Grant N0014-90-31863, ARO Grant DAAL03-89-C-0031, and NSF Grant IRI90-16592. We thank Aravind Joshi for his support for this work. We also thank Evan Antworth, Mark Foster, Lauri Karttunen, Mark Liberman, and Annie Zaenen for their help and suggestions.
[†]Visiting from Stanford University.

[1]We refer the reader to Karttunen and Wittenburg (1983) or Antworth (1990) for more details on the morphological rules.
[2]Proper nouns were not included in the tables.

| Category | # Root Forms | # Inflected Forms |
|---|---|---|
| Pronoun | 92 | 93 |
| Preposition | 148 | 150 |
| Determiner | 100 | 100 |
| Conjunction | 64 | 64 |
| Adverb | 6992 | 7176 |
| Noun | 50370 | 199303 |
| Adjective | 20550 | 65146 |
| Verb | 11880 | 45445 |
| TOTAL | 90196 | 317477 |

Figure 1: Size of the PC-KIMMO Lexicons.

## 2.1 Adjectives

The continuation classes for adjective specify that the word can undergo the rules of comparative and superlative. For example, the lexicon entry for the adjective 'funky' is:

```
funky    A_Root2   "A(funky)"
```

The entry consists of a word funky, followed by the continuation class A_Root2, and a parse "A(funky)". The continuation class specifies that the word can undergo the normal rules of comparative and superlative, and the parse states that the word is an adjective with root 'funky'. The following is a sample run of PC-KIMMO's recognizer:

```
recognizer≫funky
  funky              A(funky)
recognizer≫funkier
  funky+er           A(funky) COMP
recognizer≫funkiest
  funky+est          A(funky) SUPER
```

The output line contains the root form and any affixes, separated by '+'s. Thus, a '+' in the output indicates a morphological rule was used; its absence means no rule was used, and the parse was returned as found in the lexicon. PC-KIMMO will automatically add attributes such as COMP and SUPER to the parse, depending on the morphological rule matched by the surface form. But for irregularly inflected forms, special continuation classes indicate that the complete parse (viz., part of speech, root, and attributes) should be taken 'as is' from the lexicon entry. For example:

```
better   A_Root1   "A(good) COMP"
best     A_Root1   "A(good) SUPER"
good     A_Root1   "A(good)"
```

The class A_Root1 tells PC-KIMMO not to apply the morphological rules to 'better', 'best', and 'good'. Thus, 'gooder' is *not* recognized as 'good+er'.

```
recognizer≫best
  best               N(best) SG
  best               A(good) SUPER
  best               Adv(best)
recognizer≫good
  good               N(good) SG
  good               A(good)
recognizer≫better
  better             N(better) SG
  better             A(good) COMP
  better             V(better) INF
  better             Adv(better)
recognizer≫gooder
  *** NONE ***
recognizer≫goodest
  *** NONE ***
```

The attributes (such as COMP) can later be translated into feature structures with the help of templates as in PATR (Shieber, 1986). The list of attributes is found in Appendix A.

## 2.2 Nouns

Inflections of nouns, such as the formation of plural and genitive, are handled by morphological rules (unless the formation is idiosyncratic). In the lexicon for nouns, the continuation class N_Root1 indicates that the formation of genitive applies regularly and that no other inflection applies. The continuation class N_Root2 indicates that the formation of the plural and of the genitive apply regularly.

```
mice        N_Root1   "N(mouse) PL"
mouse       N_Root1   "N(mouse) SG"
ambassador  N_Root2   "N(ambassador)"
```

Thus, the above lexicon entries are recognized as below:

```
recognizer≫mice
  mice               N(mouse) PL
recognizer≫mouse
  mouse              N(mouse) SG
  mouse              V(mouse) INF
recognizer≫mouses
  mouse+s            V(mouse) 3SG PRES
recognizer≫mice's
  mice+'s            N(mouse) PL GEN
recognizer≫mouses'
  *** NONE ***
recognizer≫mouse's
  mouse+'s           N(mouse) SG GEN
recognizer≫ambassadors
  ambassador+s       N(ambassador) PL
recognizer≫ambassador's
  ambassador+'s      N(ambassador) SG GEN
recognizer≫ambassadors'
  ambassador+s+'s    N(ambassador) PL GEN
```

## 2.3 Verbs

Given the infinitive form of a verb, the formation of the third person singular (+s), its past tense (+ed), its past participle (+ed), and its progressive form (+ing) is handled by morphological rules unless lexical idiosyncrasies apply. In order to encode all possible idiosyncrasies over the three verb endings, eight continuation classes are defined (see Figure 2). Each continuation class specifies the inflectional rules which can apply to the given lexical item.

| Continuation class | Applicable rules |
| --- | --- |
| V_Root1 | none |
| V_Root2 | +ed |
| V_Root3 | +s |
| V_Root4 | +s, +ed |
| V_Root5 | +ing |
| V_Root6 | +ing, +ed |
| V_Root7 | +ing, +s |
| V_Root8 | +ing, +s, +ed |

Figure 2: Continuation classes for verbs

Examples of lexical entries for verbs follow:

```
admire      V_Root8    "V(admire)"
dyeing      V_Root1    "V(dye) PROG"
dye         V_Root4    "V(dye)"
zigzagging  V_Root1    "V(zigzag) PROG"
zigzagged   V_Root1    "V(zigzag) PAST WK"
zigzagged   V_Root1    "V(zigzag) PPART WK"
zigzag      V_Root3    "V(zigzag)"
tangoes     V_Root1    "V(tango) 3SG PRES"
tango       V_Root6    "V(tango)"
taught      V_Root1    "V(teach) PAST STR"
taught      V_Root1    "V(teach) PPART STR"
teach       V_Root7    "V(teach)"
```

Examples of runs follow:

```
recognizer≫admires
    admire+s         V(admire) 3SG PRES
recognizer≫admired
    admire+ed        V(admire) PAST WK
    admire+ed        V(admire) PPART WK
recognizer≫admiring
    admire+ing       V(admire) PROG
recognizer≫admire
    admire           V(admire) INF
recognizer≫dyed
    dye+ed           V(dye) PAST WK
    dye+ed           V(dye) PPART WK
recognizer≫dyes
    dye+s            N(dye) PL
    dye+s            V(dye) 3SG PRES
recognizer≫teaches
    teach+s          V(teach) 3SG PRES
recognizer≫teached
    *** NONE ***
recognizer≫taught
    taught           V(teach) PAST STR
    taught           V(teach) PPART STR
recognizer≫tangoed
    tango+ed         V(tango) PAST WK
    tango+ed         V(tango) PPART WK
recognizer≫tangoing
    tango+ing        V(tango) PROG
recognizer≫tangoes
    tangoes          V(tango) 3SG PRES
```

The attributes `WK` (for "weak") and `STR` (for "strong") mark whether the verb forms its past tense regularly or irregularly, respectively. The distinction enables unambiguous reference to homographs—words spelled identically but with different semantic and syntactic properties. For example, the verb 'lie' with the meaning 'to make an untrue statement' and the verb 'lie' with the meaning 'to be prostrate' have different syntactic and morphological behavior: the first one is regular, while the second one is irregular:

```
He has lain on the floor.
He has lied about everything.
```

Usually, it suffices to index the syntactic properties of each verb by its root form alone. However, homographs require addition information. In English, the attributes `WK` and `STR` are sufficient to distinguish homographs with different morphological behavior.

```
recognizer≫lied
    lied             N(lied) SG
    lie+ed           V(lie) PAST WK
    lie+ed           V(lie) PPART WK
recognizer≫lain
    lain             V(lie) PPART STR
recognizer≫lay
    lay              V(lay) INF
    lay              V(lie) PAST STR
```

## 2.4 Other Parts of Speech

Pronouns, prepositions, determiners, conjunctions, and adverbs are given continuation classes that inhibit the application of morphological rules. All of the morphological information is stored in the parse in the lexicon entry:

```
herself   Pron   "Pron(herself) REFL FEM 3SG"
it        Pron   "Pron(it) NEUT 3SG NOMACC"
behind    Prep   "Prep(behind)"
coolly    Adv    "Adv(coolly)"
```

PC-KIMMO recognizes them as follows:

```
recognizer≫herself
    herself          Pron(herself) REFL FEM 3SG
recognizer≫it
    it               N(it) SG
    it               Pron(it) NEUT 3SG NOMACC
recognizer≫behind
    behind           N(behind) SG
    behind           Adv(behind)
    behind           Prep(behind)
recognizer≫coolly
    coolly           Adv(coolly)
```

# 3 Lexicons as a Database

PC-KIMMO builds in memory a data structure from the complete lexicon. Consequently, our large lexicons occupy more than 19 Mbytes of process memory. Further, the large size of the structure implies long search times as PC-KIMMO swaps pages in and out.

Thus, to solve both the time and space problems simultaneously, we compiled all inflectional forms into a disk-based database using a UNIX hash table facility (Seltzer and Yigit, 1991).

To compile the database, we used PC-KIMMO as a generator, inputting each root form and all the endings that it could take, as indicated by the continuation class. The resulting inflected form became the key, and the associated morphological information was then inserted into the database.

For example, the PC-KIMMO lexicon file contains the entry:

```
saw    N_Root2   "N(saw)"
```

The class `N_Root2` indicates that the noun 'saw' forms its plural, singular genitive, and plural genitive regularly. Thus, we send to the generator three lexical forms and the three suffixes for each inflection, extracting three inflected surface forms:

```
Lexical   saw+s    saw+'s    saw+s+'s
Surface   saws     saw's     saws'
```

The root form of a noun is identical with the singular inflection, so we have a total of four inflected forms. Since we know which suffix we added to the root, we also know the attributes for that inflection. The inflected form becomes the key, while the part of speech, root, and attributes are stored as the content in the database. Hence, the lexicon entry for the noun 'saw' produces four key–content pairs in the database: (saw, saw N SG), (saws, saw N PL), (saw's, saw N SG GEN), (saws', saw N PL GEN).

Likewise, the verb lexicon contains the entries:

```
saw    V_Root8   "V(saw)"
saw    V_Root1   "V(see) PAST STR"
```

The continuation class `V_Root8` indicates four inflections besides the infinitive: third-person singular (+s), past (+ed), weak past participle (+ed), and present participle (+ing). Hence, the generator produces:

```
Lexical   saw+s    saw+ed    saw+ing
Surface   saws     sawed     sawing
```

The class `V_Root1` allows no inflections, but builds the inflection–feature pair directly: (saw, see V PAST STR).

Hence, morphological analysis is reduced to sending the surface forms to the database as keys and retrieving the returned strings. Figure 3 lists the database keys and content strings produced by the three lexicon lines given above. Note that distinct entries are separated by '#'. Since multiple lexical forms can map to the same surface form, the actual number of keys (*ca.* 292000) is less than the number of lexical forms (*ca.* 317000). Also, with the database residing on the disk, access times average 6 to 10 milliseconds, which greatly improves upon PC-KIMMO.

### 3.1 Implementation Considerations

The large number of keys implies a very large disk file. To reduce the size of the file, we take advantage of the morphological similarity in English between an inflected form and its lexical root form. Indeed, the root is often contained intact within the inflected form.

| Key | Contents |
|---|---|
| saw | saw N SG#saw V INF#see V PAST STR |
| saws | saw N PL#saw V 3SG PRES |
| saw's | saw N SG GEN |
| sawing | saw V PROG |
| sawed | saw V PAST WK#saw V PPART WK |
| saws' | saw N PL GEN |

Figure 3: Database pairs

Hence, instead of storing the root, we store the number of shared characters along with any differing characters, and reassemble the root from the inflected form on each database query. Further, despite the large set of attributes, relatively few combinations (*ca.* 80) are meaningful, and can be encoded in a single byte. Since a large proportion of roots are wholly contained within the surface form, and since 92% of the keys have one lexical entry, the average content string is only three bytes long. Consequently, the total disk file is under 9Mbytes. We anticipate further compaction in the near future.

### 3.2 Accompanying Utilities

Besides the PC-KIMMO lexicons, we currently maintain the database file and an ASCII-character "flat" version for on-line database browsing. One program converts the lexicons into the database format, while others dump the database into the flat file or reconstruct the database from the flat file. We have also built a X Windows tool to perform maintenance on the database file (see Figure 4). This tool automatically maintains the consistency between the flat file and the database file. We have built hooks in C and Lisp (Lucid 4.0) to access either the database or PC-KIMMO from within a running process.

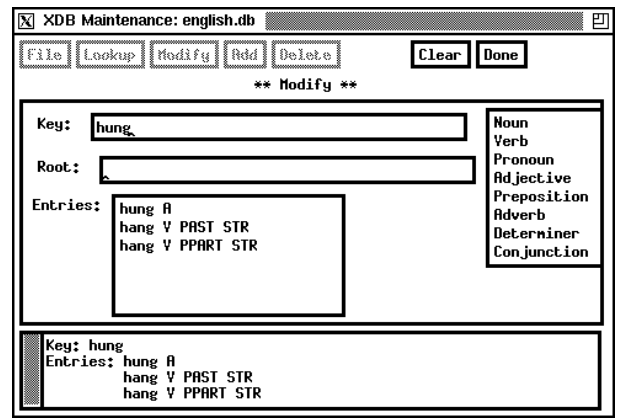

Figure 4: Morphological Database X Window Tool

## 4 Obtaining the Analyzer

The PC-KIMMO lexicons, the database files, the LISP and C access functions, programs for converting between formats, and the X Window maintenance tool are available without charge for research purposes. Please send e-mail to `lex-request@linc.cis.upenn.edu`.

## 5 Conclusion

We have presented freely available morphological tables and a morphological analyzer to handle English inflections. The tables handle approximately 317000 inflected forms corresponding to 90000 stems.

These tables can be used by an implementation of a two-level processor for morphological analysis such as PC-KIMMO.

However, these large tables degrade the performance of PC-KIMMO's current implementation, requiring about 18 Mbytes of RAM while slowing the access time.

To overcome these shortcomings, we created a morphological analyzer consisting of a disk-based database using a UNIX hash table facility. With this database, access times average 6 to 10 milliseconds while moving all of the data to the disk. We also provide an X Window tool for facilitating the maintenance and access to the database.

The package is ready to be integrated into an application such as a parser. Hooks written in Lisp and C for accessing these tables are provided.

To our knowledge, this package is the only available free English morphological analyzer with very wide coverage.

## A List of Attributes

| | |
|---|---|
| 1SG | 1st person singular |
| 2SG | 2nd person singular |
| 3SG | 3rd person singular |
| 1PL | 1st person plural |
| 2PL | 2nd person plural |
| 3PL | 3rd person singular |
| 2ND | 2nd person |
| 3RD | 3rd person |
| SG | singular |
| PL | plural |
| PROG | progressive |
| PAST | past tense |
| PPART | past participle |
| INF | infinitive or present (not 3rd person) |
| PRES | present |
| STR | strongly inflected verb |
| WK | weakly inflected verb |
| GEN | genitive (+ 's) |
| NOM | nominative case |
| ACC | accusative case |
| NOMACC | nominative or accusative case |
| NEG | negation |
| PASSIVE | passive form (for "born") |
| to | contracted form verb + to |
| COMP | comparative |
| SUPER | superlative |
| MASC | masculine |
| FEM | feminine |
| NEUT | neuter |
| WH | wh-word |
| REFL | reflexive |
| REF1SG | 1st person singular referent |
| REF2ND | 2nd person referent |
| REF2SG | 2nd person singular referent |
| REF2PL | 2nd person plural referent |
| REF3SG | 3rd person singular referent |
| REF3PL | 3rd person plural referent |
| REFMASC | masculine referent |
| REFFEM | feminine referent |